\documentclass[letterpaper, 10 pt, journal, twoside]{IEEEtran}
\usepackage{helvet}         		
\usepackage{type1cm}        		
\usepackage{graphics} 		
\usepackage{graphicx}        
\usepackage{epsfig} 			
\usepackage{multicol}        		
\usepackage[bottom]{footmisc}	

\usepackage{amssymb}
\usepackage[ruled]{algorithm2e}
\usepackage{subfigure}
\usepackage{newtxtext}       %
\usepackage{newtxmath}       
\usepackage{caption}
\usepackage{gensymb}
\usepackage{booktabs}
\usepackage[table,xcdraw]{xcolor}
\usepackage{epstopdf}
\usepackage{url}
\usepackage[font={small}]{caption}
\usepackage{commath}
\usepackage{subfigure}
\usepackage{multirow}
\usepackage{makecell}
\usepackage{amsmath}
\usepackage{arydshln} 
\usepackage{hyperref}

\ifCLASSINFOpdf
\else
\fi
\newcommand{\Tbf}{\mathbf{T}}

\newcommand{\Ical}{\mathcal{I}}

\begin{document}

\title{Above and Below: Heterogeneous Multi-robot SLAM Across Surface and Underwater Domains}


\author{John McConnell$^{1}$, Armon Shariati$^{2}$, Paul Szenher$^{3}$ and Yaxuan Li$^{3}$\vspace{-6mm}
\thanks{$^{1}$J. McConnell is with the United States Naval Academy, Annapolis, MD, USA. {\tt\small jmcconne@usna.edu}. $^{2}$A. Shariati is with Shield-AI, San Diego CA, USA. {\tt\small armon.shariati@shield.ai}. $^{3}$P. Szenher and $^{3}$Y. Li are with the Stevens Institute of Technology, Hoboken, NJ, USA. {\tt\small$\{$pszenher$,$yli21$\}$@stevens.edu}.}
}


%



\maketitle

\begin{abstract}
Multi-robot simultaneous localization and mapping (SLAM) is a fundamental task in multi-robot operations. Robots must have a common understanding of their location and that of their team members to complete coordinated actions. However, multi-robot SLAM between Uncrewed Surface Vessels (USVs) and Autonomous Underwater Vehicles (AUVs) has primarily been achieved through acoustic pinging between robots to retrieve range measurements; a measurement technique requires that robots to be in similar locations simultaneously, have an uninterrupted path for signal propagation, and may necessitate synchronized clocks. This is especially challenging in complex, cluttered maritime environments, where structures may impede signals. However, these same structures may be observable above and below the water's surface, presenting an opportunity for inter-robot SLAM loop closure between USV and AUV data streams. This work builds upon recent research on inter-robot SLAM loop closure between USV and AUV data \cite{msm}, extending it to propose a centralized multi-robot SLAM system. Each robot performs its state estimation, and we detect loop closures between each AUV and the USV data. These inter-robot loop closures are used to merge each robot's state estimate into a centralized graph, yielding estimates for the whole time history of the USV and all AUVs in the system. Validation is performed using real-world perceptual data in three different environments. Results show improved errors for AUVs in the multi-robot SLAM system compared to single-robot SLAM over the same trajectories. To our knowledge, this is the first instance of a multi-robot SLAM system with AUVs and USVs built on loop closures rather than acoustic distance measurements. The views expressed in this document are those of the author(s) and do not reflect the official policy or position of the U.S. Naval Academy, Department of the Navy, the Department of Defense, or the U.S. Government. 
\end{abstract}

\vspace{-2mm}

%
\IEEEpeerreviewmaketitle

\vspace{-4mm}
\section{Introduction} \vspace{-2mm}
\IEEEPARstart{A}{utonomous} underwater vehicles (AUVs) have become critical tools in many settings, including Oil \& Gas, aquaculture, offshore renewable energy, and defense. AUVs are routinely employed for inspection, repair, and tasks where using human divers might be dangerous or otherwise impossible \cite{SLB-2020}. However, underwater localization is challenging compared to other domains due to the lack of GPS and low-quality perceptual data. 

A notable approach to addressing the difficulties of AUV localization is a multi-robot solution. Uncrewed Surface Vessels (USV) have easy access to high-accuracy localization through GPS or high-quality perceptual systems. In a multi-robot setting, USV localization is leveraged to enhance the quality and robustness of AUV localization. Multi-robot localization typically comes in two forms, \textit{direct} and \textit{indirect} encounters \cite{paull-2015,paull-2014}. Direct encounters typically involve acoustic ranging between robots. However, there are limitations to direct encounters, mainly requiring proximity between robots, a clear line of sight for signal propagation, and \textit{may} require robots to have synchronized clocks. Indirect encounters associate targets observed in the environment to estimate measurements between robots, offering several significant advantages: they do not need robots to be simultaneously in the same part of the environment, can be resolved whenever communication is available, and do not require clock synchronization. 

In this work, we consider teams of USVs and AUVs operating in complex, cluttered maritime environments, where underwater structures frequently impede the path of acoustic ranging and communication. However, we consider the utility of human-made structures in littoral settings. Such structures are often observable from above and below the water’s surface, including bridges, docks, ships, and pier pylons. This work will leverage the insight that common structures are observable above and below the water’s surface and use this to develop indirect encounters between USVs and AUVs. Therefore, we consider two important sensing models: (1) wide aperture multi-beam imaging sonar for its robustness to water quality, immunity to lighting conditions, and long sensing range. (2) 3D above water LiDAR for its dense, high-quality information, making it simple to render observations \textit{near} the waterline regardless of USV roll and pitch. This paper utilizes recent developments in indirect encounters (loop closure) between AUV and USV data streams \cite{msm}. However, unlike previous work, we use indirect encounters in a fully fledged multi-robot simultaneous localization and mapping (SLAM) system. 

\begin{figure}[t]
\centering
{\includegraphics[width = .999\linewidth]{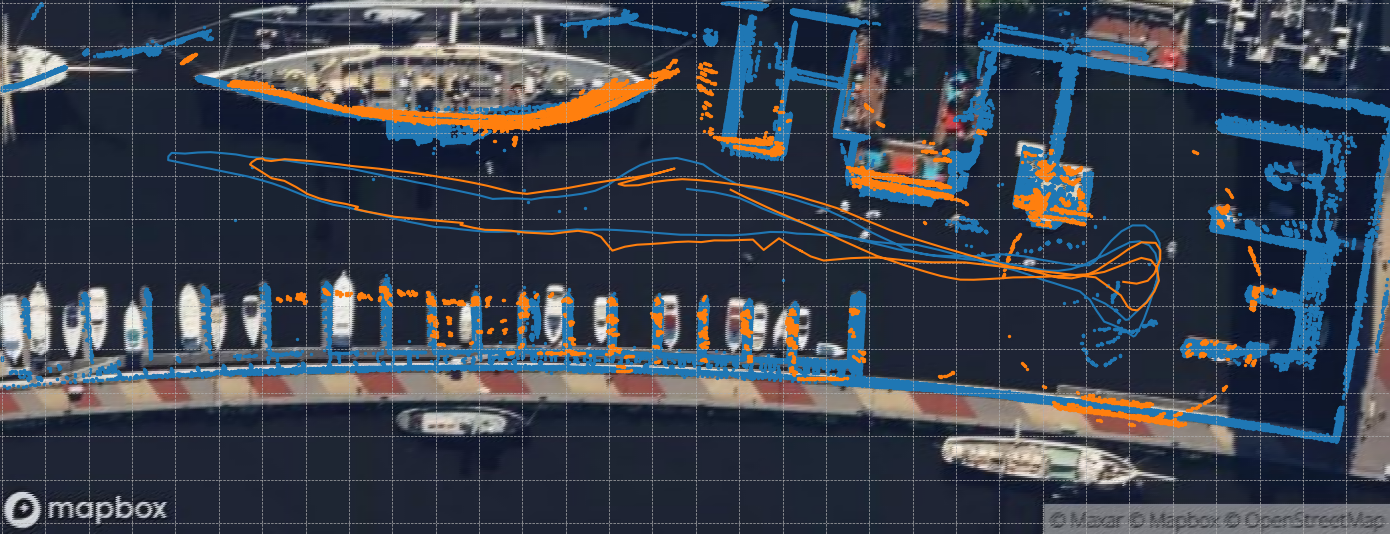}}
\caption{\textbf{Multi Robot Map Result.}  A sample multi-robot mission from the harbor environment. The blue points show the 2D LiDAR map, with the blue line showing the USV trajectory. Orange points indicate the sonar map, with the orange line representing the AUV's trajectory. Grid lines are shown on the satellite image at 10-meter cell size. }
\label{fig:lead_off}
\vspace{-5mm}
\end{figure}

Our contributions are as follows:
\vspace{-1mm}
\begin{itemize}
    \item The first example of multi-robot SLAM for USVs and AUVs using indirect encounters from perceptual data. \textcolor{black}{The proposed system's novelty is the improvement and integration of \cite{msm} into the broader SLAM problem. }
    \item A compression method for sonar data \textcolor{black}{that maintains cell resolution}, making perceptual data exchange amenable with low bandwidth communications systems.
    \item Validation of the system in three real-world environments \textcolor{black}{using simulated communications}. Showing improved AUV localization quality over single robot SLAM. \textcolor{black}{To the best of our knowledge, these are the first experiments of this type and scale that include ground truth.} \textcolor{black}{This is the most important, and broadest contribution of this work. }
    \item An open-source repository for the code and datasets used in this paper. \href{https://github.com/Jake-maritime-lab/Above-and-Below-SLAM/}{Code Link.}
\end{itemize}
\vspace{-6mm}
\section{Related Work}
\subsection{Above Water Prior Information}
Related to our work is the use of prior information above the water's surface to assist in the underwater SLAM problem. \cite{santos_1} develops a similarity scoring system between an underwater sonar image and a sector of a satellite image of the same sensor footprint using deep learning. In follow-up work, the same authors extend this similarity score to serve as the backbone of a particle filter localization system for an underwater robot equipped with imaging sonar and a priori satellite images \cite{santos_2}. Lastly, \cite{overhead-mcconnell} provides an example of sonar-based pose graph SLAM, incorporating constraints from a satellite image prior and standard place recognitions. 
\subsection{Place Recognition}
A fundamental task in robotics is place recognition, which involves detecting when a robot is in a location it has visited before and finding transformations between past and current timesteps. In the LiDAR sensing domain, Scan Context \cite{context,intensity-context,context-plus} has demonstrated impressive results without the need for machine learning. Scan context converts dense, high-resolution 3D point clouds from LiDAR into coarse context images. Scan context computes single-dimension descriptors to generate place recognition candidates using a KD-tree \cite{kd-tree}. Scan context has recently been applied in underwater imaging sonar \cite{sonar_context} and is an essential module in DRACo-SLAM \cite{draco}, a multi-robot SLAM system designed for underwater robots equipped with imaging sonar. Object scene graphs have also been studied for imaging sonar place recognition \cite{santos-2018}, requiring a high number of objects in view to compute a transform. Machine learning is used in side-scan sonar place recognition \cite{ribeiro-2018,larsson-2020}; however, there is little publicly available underwater data, making it challenging to apply these methods to new problem settings. Lastly, \cite{Wang-2021} applies the iterative closest point (ICP) algorithm \cite{icp}, using dead reckoning as an initial guess, to evaluate place recognition.

It is crucial to note that a single erroneous place recognition can lead to highly inaccurate robot pose estimation; for this reason, we consider robust outlier rejection systems. Pairwise consistent measurement set maximization (PCM) \cite{pcm} evaluates measurements as a set rather than in isolation, acting as a pre-filter to the SLAM backend. Dynamic covariance scaling (DCS) \cite{dcs} adjusts measurement covariance during the solver step, and switchable constraints \cite{switch} turn measurements on or off. While these robust backends are impressive, they often defer to the odometry chain, which, in the case of multi-robot SLAM, could be challenging given that the measurements between robots may be sparse. 

\subsection{Cooperative Localization}
Measurements between robots typically come in two forms: direct and indirect encounters \cite{paull-2015}. Direct encounters occur when robots measure the distance between them, often achieved through acoustic pinging. Indirect measurements are developed by associating commonly observed targets in the environment to form transformations between robots \cite{paull-2014}. A significant drawback of relying solely on direct encounters is that robots must be in the same part of the environment, with an unbroken path for the signal to pass between them. Moreover, while many methods avoid the need for clock synchronization \cite{no_sync_1}\cite{no_sync_2}, many others require it to develop range measurements between robots {\cite{sync_1,sync_2,sync_3}}. 

There are several notable examples of collaborative localization using direct encounters. Firstly, \cite{fischell-2019} shows a team of underwater robots using direct encounters with a surface station to perform collaborative localization. \cite{yao-2009} shows an example of a pair of robots operating in a similar pattern. \cite{Halsted} shows a collaborative localization method with distributed processing. \cite{li-2019} introduces collaborative localization using several fixed-position beacons, comparing the performance across a robot team with and without the addition of fixed beacons. \cite{bahr-2009} evaluates several system variations that include onboard dead reckoning systems and inter-robot pinging. When considering work with indirect encounters, \textcolor{black}{the above water setting has shown numerous impressive contributions \cite{disco}.} In a maritime setting \cite{paull-2015} presents a simulated example of an underwater multi-robot team using GPS at the surface, inter-robot pinging, and a data association method that relies on GPS for initialization. DRACo-SLAM \cite{draco} demonstrates an example of multi-robot SLAM using underwater robots with imaging sonar and no direct encounters. \cite{msm} finds inter-robot loop closures between underwater imaging sonar and above-water 3D LiDAR; however, this work focuses on the loop closure problem and does not integrate this tool with a SLAM solution, \textcolor{black}{nor does it consider the cost of data exchange. }

\subsection{\textcolor{black}{Sonar Data Compression}}
\textcolor{black}{When considering transmitting sonar data, compression is crucial due to bandwidth limitations.}
\textcolor{black}{DRACo-SLAM \cite{draco} compresses data using a voxel grid, degrading resolution.}
\textcolor{black}{{\cite{sonar-compression-dnn} proposes a machine learning approach to sonar image compression; however, new datasets may be out of distribution, potentially degrading performance.} In this work, we propose a simple method that we are confident will generalize beyond the experiments presented in this paper. }


\begin{figure*}[t]
\centering
{\includegraphics[width = .65\linewidth]{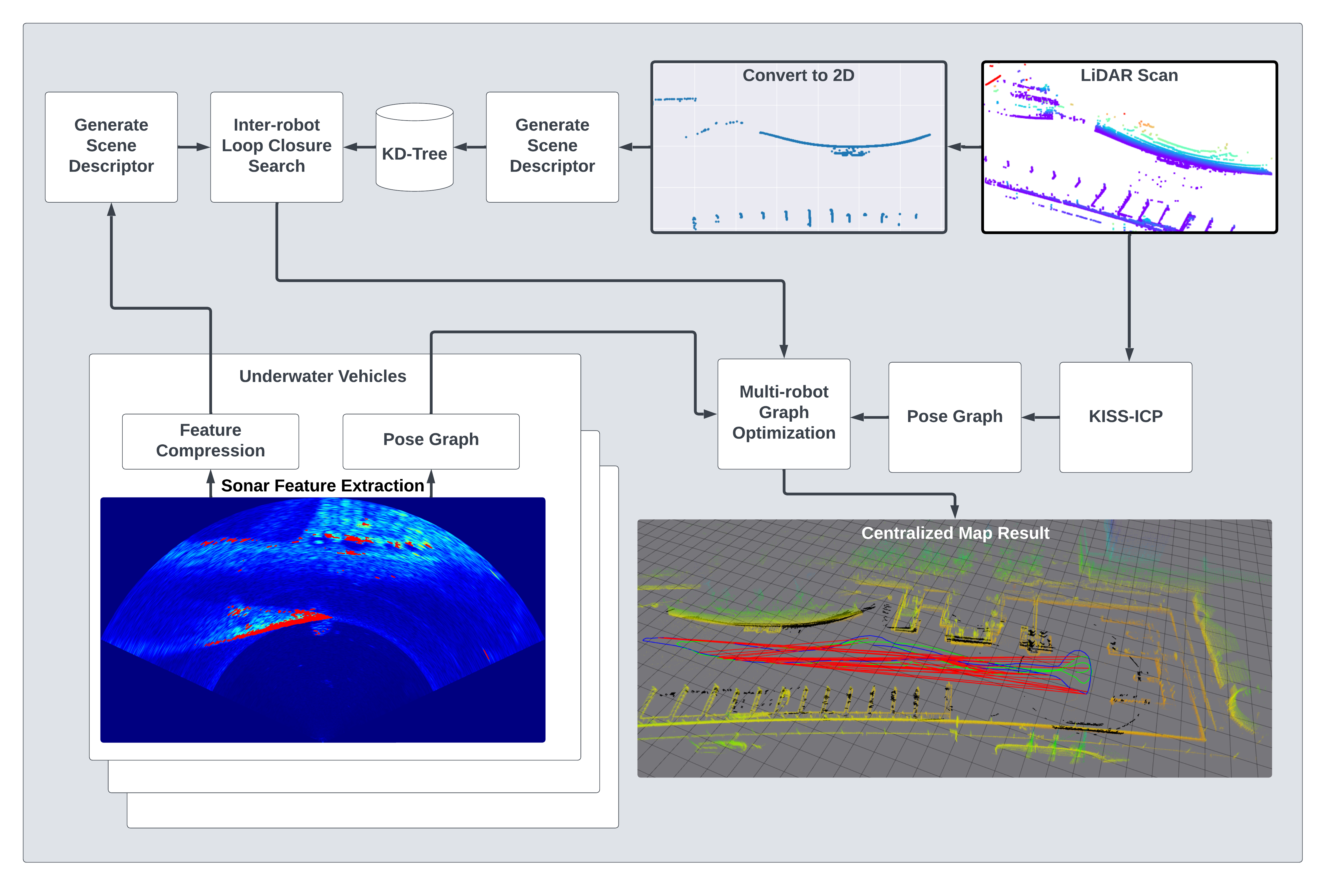}}
\caption{\textbf{System Flow.} At bottom left, we show $N$ underwater vehicles with imaging sonar; each AUV compresses its features and computes a pose graph, sending results to the USV. The USV converts its LiDAR scans into 2D scans near the water's surface. The USV performs a loop closure search between the AUV's sonar data and the USV's LiDAR data. Loop closures are used to merge USV and AUV pose graphs, resulting in a centralized state estimate and a joint map, as shown in the bottom right. 3D LiDAR point clouds are shown, with color mapped to height. Black points represent the sonar map, the blue line indicates the USV trajectory, the green line represents one of the AUV trajectories, and the red lines denote inter-robot loop closures. }
\label{fig:flow_chart}
\vspace{-5mm}
\end{figure*}

\section{Problem Definition}
In this work we consider a team of $N$ robots.   
Each robot $n \in N$ maintains its own state estimate, in its own reference frame $\Ical_n$. 
Each robot collects observations, $\mathbf z_{n,t}$  
at pose $\mathbf x_{n,t}$ at time step $t$. Each robot pose is defined in the plane as
\vspace{-2mm}
\begin{align}
    \mathbf x_{n,t}
    = \begin{pmatrix} x_r\textcolor{black}{,} \ 
    y_r\textcolor{black}{,} \
    \theta_r \end{pmatrix}^\top, 
\end{align}
and each set of observations is a two dimensional point cloud. \textcolor{black}{AUVs use sonar, with returns in spherical coordinates denoted as ranges $r_s\in \mathbb{R}_+$, bearings $\theta_s \subseteq [-\pi,\pi)$, elevation angles $\phi_s \subseteq [-\pi,\pi)$, and associated intensity values $\gamma_s \in \mathbb{R}_+$. Imaging sonar however, does not measure $\phi_s$, and therefore we treat it as zero, or in-plane. USVs use LiDAR, with returns in spherical coordinates denoted as ranges $r_L\in \mathbb{R}_+$, bearings $\theta_L \subseteq [-\pi,\pi)$, elevation angles $\phi_L \subseteq [-\pi,\pi)$, and associated intensity values $\gamma_L \in \mathbb{R}_+$} Each robot moves through the environment according to the dynamics
\vspace{-3mm}
\begin{align}
    \mathbf x_{n,t} = \mathbf g(\mathbf u_{n,t} , \mathbf x_{n,t-1}) + \textcolor{black}{\epsilon_{n,t}},
\end{align}
where $\mathbf x_{n,t-1}$ is the previous timestep’s pose, $\mathbf u_{n,t}$ is the control command and $\textcolor{black}{\epsilon_{n,t}}$ is process noise. The posterior probability over the time history of poses is defined as
\vspace{-3mm}
\begin{align}
    p(\mathbf x_{n,1:t} , \mathbf m_n | \mathbf z_{n,1:t}, \mathbf u_{n,1:t}  ),
    \label{eq:post}
\end{align}
with map $\mathbf m_n$. Lastly, while we propose a multi-robot SLAM solution, there is no notion of initial transform between reference frames $\Ical_n$. \textcolor{black}{We assume robots may communicate with one another for the entire mission duration, and only do so at discrete SLAM steps. }

\section{Algorithm}
\subsection{USV State Estimation}
While there are many methods for USV state estimation, we utilize LiDAR odometry. We found it to be simple, effective, \textcolor{black}{provides heading, does not require GPS, yields much higher accuracy than GPS }and since the downstream processes use LiDAR, this sensor is present onboard the USV. However, we would like to emphasize that \textbf{any} localization system could be used here.

\subsubsection{LiDAR Pre-processing}
\label{lidar-odom-pre-processing}
Maritime environments can introduce noise into LiDAR point clouds due to reflections off the water surface. These reflections can even mirror the above-water structures below the waterline. For this reason, we form a partial state estimate for the USV, 

\vspace{-7mm}
\begin{align}
    \label{partial_state}
    \mathbf{x_{t,p}} = \ \begin{bmatrix}\ z_\mathcal{I},  \ \phi_\mathcal{I}, \ \psi_\mathcal{I} \end{bmatrix}^T.
\end{align}
Where $z_\mathcal{I}$ is the height of the LiDAR, measured from the water surface, $\phi_\mathcal{I}$ is the roll, and $\psi_\mathcal{I}$ is the pitch. $z_\mathcal{I} $ is sourced from calibrating the height of the LiDAR above the water's surface using a depth sensor mounted below the waterline with $\phi_\mathcal{I}$ and $\psi_\mathcal{I}$ from the IMU. The LiDAR scan $\mathcal{P}_L$ is corrected for height, roll, and pitch using this partial state estimate. $\mathcal{P}_L$ is cropped according to height, where only points above the waterline are retained, denoted as $\mathcal{\hat{P}}_L$. This method is simple yet effective for removing erroneous information from LiDAR point clouds. 

\subsubsection{KISS-ICP}
\label{kiss-icp}
We select KISS-ICP \cite{kiss-icp} for its simplicity, robustness, lack of system requirements, and functionality on this maritime data. KiSS-ICP aligns each LiDAR scan, $\mathcal{\hat{P}}_L$, to the local LiDAR map, $\mathcal{M}_L$, a composite of previous scans, and provides a pose estimate for each LiDAR timestep in the USV reference frame.

\subsubsection{Pose Graph}
\label{lidar-pose-graph}
While pose information is available at each LiDAR timestep, we downsample this information to form a more sparse representation. If the USV has moved at least $\Delta T_{min}$ or rotated $\Delta R_{min}$, we formulate a LiDAR odometry factor, $\mathbf f^{\text{LO}}$, which is the transformation from the last time discrete pose. Forming the factor graph, 

\vspace{-5mm}
\begin{flalign}
\label{multi-robot-graph}
\mathbf f(\boldsymbol \Theta) = \mathbf  f^{\text{0}}(\boldsymbol \Theta_0) & \prod_i \mathbf f^{\text{LO}}_{i}(\boldsymbol \Theta_i), 
\end{flalign}
which does not contain a source of loop closure. We found that LiDAR loop closure was unnecessary in testing, and simple LiDAR odometry provided highly accurate, stable results. However, it can easily be incorporated using similar systems in future work \cite{kiss-slam}.

\subsection{AUV State Estimation}
In this section, we consider state estimation onboard the AUV. Similar to the USV, the source of this state estimation could be any system. We consider a dead-reckoning pose graph combining Doppler velocity logger (DVL) and IMU data. 
\subsubsection{Pose Graph}
\label{auv-pose-graph}
We combine DVL+IMU information in an extended Kalman filter (EKF) \cite{kalman-filter}, yielding AUV position estimates. \textcolor{black}{Since AUVs are free floating, the Kalman filter motion model is simply particle motion.} \textcolor{black}{Due to space constraints, we direct interested readers to prior work for more details \cite{kalman-filter}.} While DVL+IMU pose information is available at a high rate, we down-sample this information to a more sparse representation. If the AUV has moved at least $\Delta T_{min}$ or rotated $\Delta R_{min}$, we create a DVL odometry factor, $\mathbf f^{\text{DO}}$, the transformation between the current and previous time discrete poses. Yielding the factor graph, 
\begin{flalign}
\label{multi-robot-graph}
\mathbf f(\boldsymbol \Theta) = \mathbf  f^{\text{0}}(\boldsymbol \Theta_0) & \prod_i \mathbf f^{\text{DO}}_{i}(\boldsymbol \Theta_i).
\end{flalign}
\textcolor{black}{All factor graphs in this paper are implemented using the GTSAM library \cite{gtsam}.} By design, this factor graph does not contain a source of loop closure; the factor graph yields a stable, yet drifting estimate to build our system off of. Downstream systems will layer in a source of loop closure, which we will compare in experiments to a more standard, single-robot approach. 

\begin{figure}[t]
\centering
\subfigure[Grid of raw contacts \label{fig:cartoon_raw}]{\includegraphics[width=.25\columnwidth]{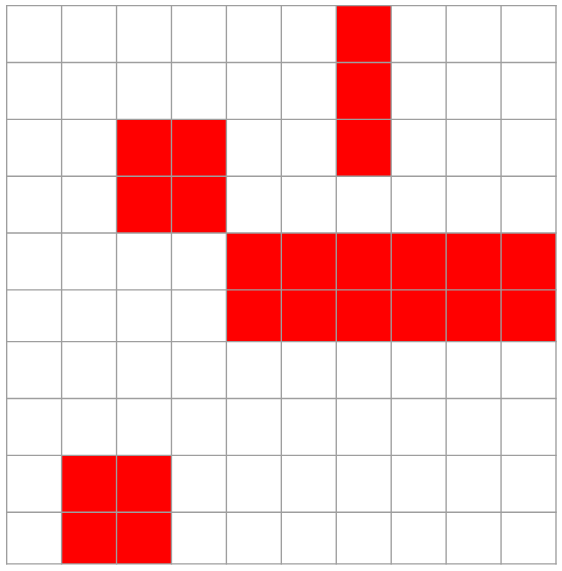}}\ \;
\subfigure[Reduction to rectangles \label{fig:cartoon_box}]{\includegraphics[width=.25\columnwidth]{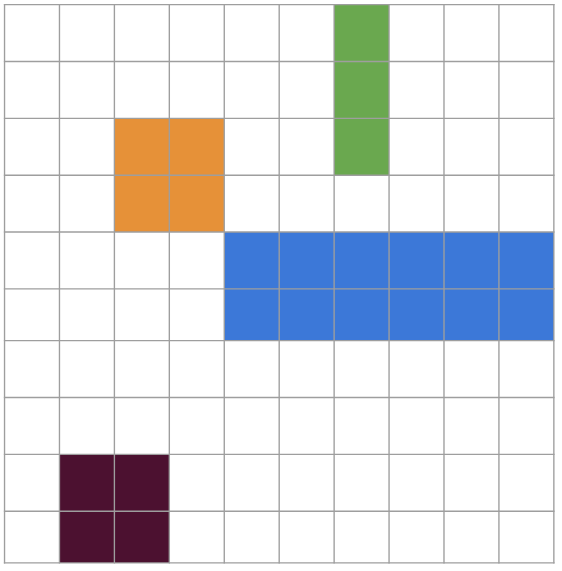}}\\
\subfigure[Rectangles drawn on real sonar image \label{fig:sonar_image_box_compression}]{\includegraphics[width=.55\columnwidth]{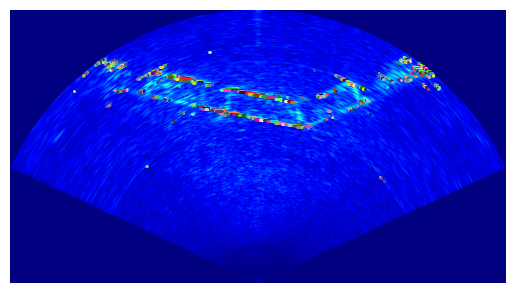}}\ \;

\caption{\textbf{Sonar image compression.} (a) shows an illustration of a sonar image with multiple contacts highlighted in red. (b) shows the results of reducing each of these patches to a rectangle represented by the top left and bottom right corners. (c) shows a real-world example sonar image of (b), where all sonar contacts have been grouped into rectangles, each with a different color.  }
\label{fig:compression}
\vspace{-5mm}
\end{figure}

\subsection{Communications Strategy}
\label{comms-strat}
In this work, we consider a centralized multi-robot SLAM system. The USV acts as the central node, solving the multi-robot SLAM problem using the information provided by AUVs. Each AUV sends a message to the USV consisting of its current pose estimate and associated sonar data at each discrete timestep. This one-way communication of perceptual data is important, as we do not communicate LiDAR point clouds from USV to AUV at any time, avoiding the need to expend limited data exchange capacity on this costly data structure. 

\begin{figure*}[th]
\centering
\subfigure[Bridge Satellite Image\label{fig:bridge_sat_image}]{\includegraphics[height=3.2cm]{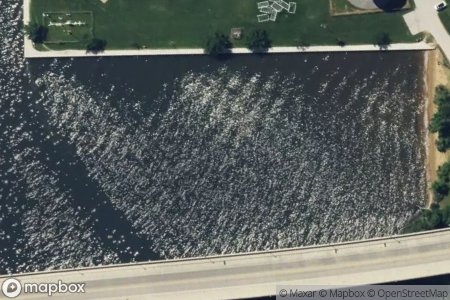}}\ \;
\centering
\subfigure[Waterfront Satellite Image\label{fig:waterfront_sat_image}]{\includegraphics[height=3.2cm]{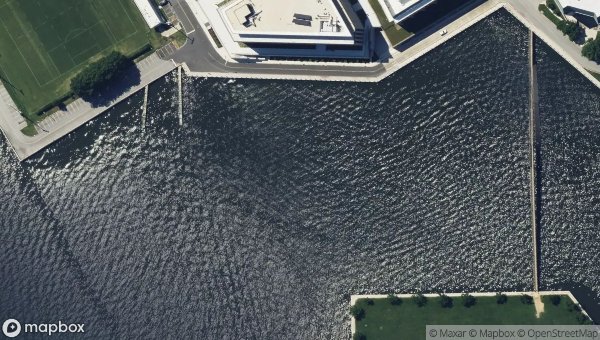}}\\
\subfigure[Harbor Satellite Image\label{fig:harbor_sat_image}]{\includegraphics[height=3.2cm]{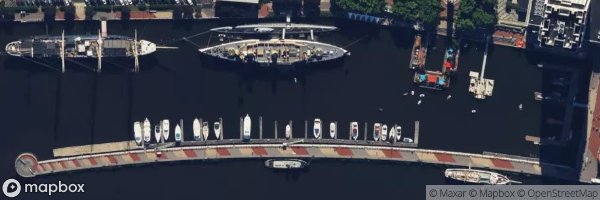}}\ \;
\subfigure[Kingfisher Robot \label{fig:kingfisher}]{\includegraphics[height=3.2cm]{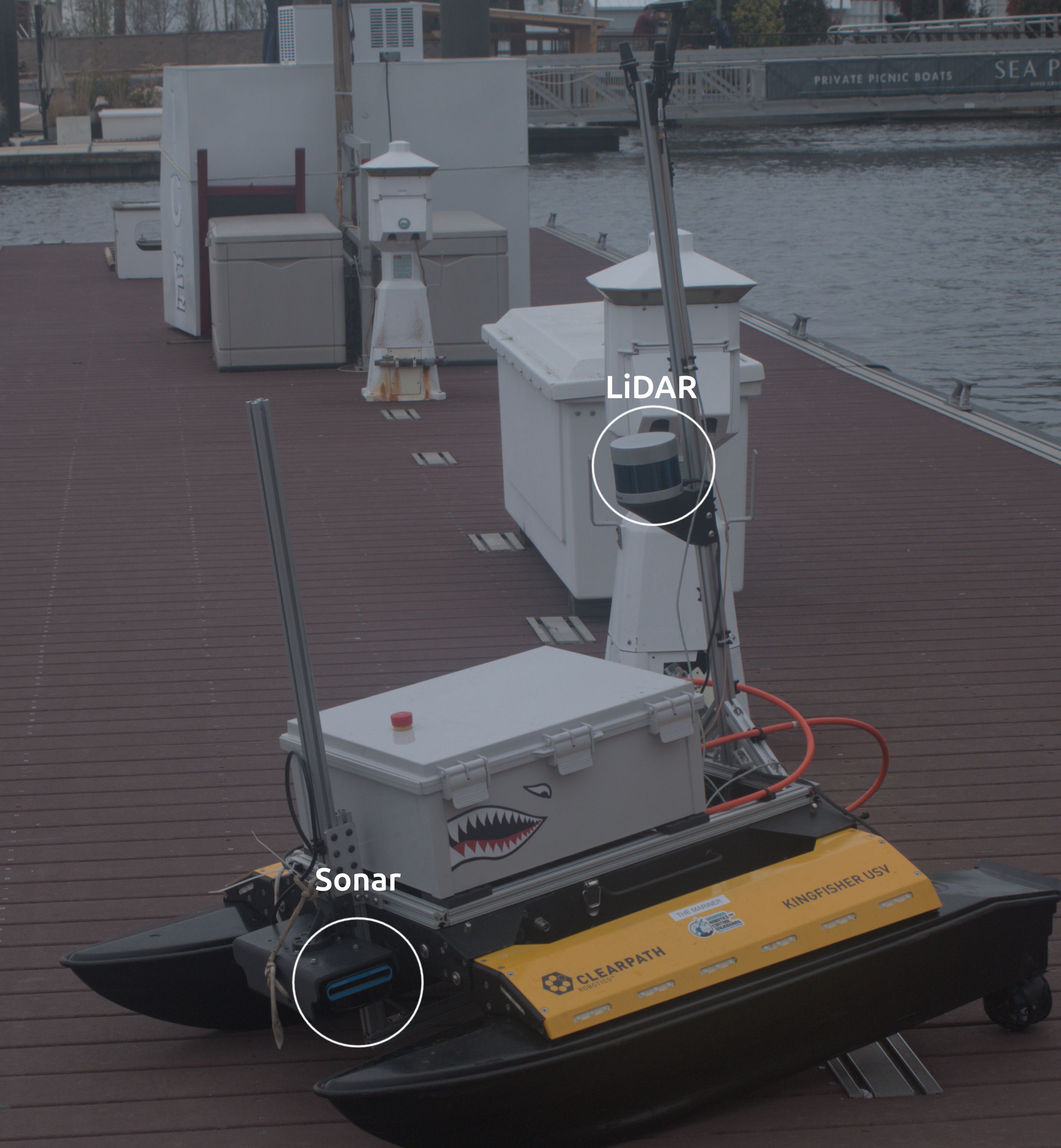}}\ \;

\caption{\textbf{Test Environments and real robot.} (a) shows the bridge, (b) shows waterfront and (c) shows harbor. (d) shows our kingfisher robot with sonar and LiDAR used to collect data for the experiments in this work. Satellite images are from Mapbox \cite{mapbox}.}
\label{fig:envs}
\vspace{-5mm}
\end{figure*}

\subsubsection{Sonar Pre-processing}
\label{sonar-cfar}

Recall that sonar data originates as polar coordinate images, which are costly to transmit. For this reason, we do not propose transmitting raw sonar images from the AUV to the USV. Instead, we select which pixels are legitimate contacts with the environment, using SOCA-CFAR \cite{cfar}, denoted as $\mathcal{P}_c$, contacts. By transmitting only the pixel coordinates of legitimate sonar contacts, represented by a pair of 32-bit integers, we achieve significant savings over raw images.


\subsubsection{Sonar Compression}
\label{sonar-compression}
To further improve communication efficiency, we apply a compression technique to the sonar contacts, $\mathcal{P}_c$, with an illustrative example shown in Figure \ref{fig:cartoon_raw}. We reduce $\mathcal{P}_c$ to rectangular patches denoted as $\mathcal{P}_r$ and illustrated in Figure \ref{fig:cartoon_box}.
Contacts that are not part of a patch are retained using their coordinates. Rather than transmitting the coordinates of all contacts as shown in Figure \ref{fig:cartoon_raw}, we can transmit the top left and bottom right coordinates of each patch shown in Figure \ref{fig:cartoon_box}, with a real-world example of these patches shown in Figure \ref{fig:sonar_image_box_compression}. While the efficiency of this compression varies with the environment, \textcolor{black}{it maintains the cell resolution, where other methods \cite{draco} reduce resolution, } and experimental results will demonstrate its effectiveness.



\subsection{Robot Loop Closure}
\subsubsection{Loop Closure LiDAR Pre-processing}
\label{lidar-2D-proccessing}
In downstream systems, our system will search for possible loop closures between USV LiDAR scans and AUV sonar scans. However, 3D LiDAR scans must be rendered into a standardized representation with sonar data, such as a 2D point cloud. This 2D point cloud consists of points \textit{near} the water's surface. In Section \ref{lidar-odom-pre-processing}, we remove erroneous points and any data presented below the water. This LiDAR scan $\mathcal{\hat{P}}_L$ is further processed by removing all points with a height above $h_{max}$, yielding LiDAR scan $\mathcal{\hat{P^*}}_L$. Finally, we drop the height axis of the LiDAR scan altogether, creating a 2D point cloud near the water's surface $\mathcal{\tilde{P}}_L$.

\subsubsection{Loop Closure Sonar Pre-processing}
Sonar data arrives onboard the USV as $\mathcal{P}_r$, which are pixel locations in a polar coordinate image, which are converted to cartesian 2D sonar point clouds, $\mathcal{P}_s$. These point clouds are in the plane, assuming a zero elevation angle. This detail is important since, with an imaging sonar, this elevation angle is unknown. However, in our work, this zero elevation angle represents a reasonable approach to registering sonar and LiDAR point clouds. 

\subsubsection{Inter-robot Loop Closure Search}
\label{loop-closure-search}
To find correspondences between USV-LiDAR and AUV-sonar data streams, we form scene descriptors to query for possible matches between scenes. While there are numerous ways to develop these descriptors, we employ a simple and effective method, namely range histograms, similar to many notable loop closure systems \cite{context,intensity-context,context-plus,sonar_context,draco}. Each 2D point cloud, $\mathcal{\tilde{P}}_L$ and $\mathcal{P}_s$ is converted from x-y coordinates, to range. We then enforce a maximum range $r_{max}$ on both data products and discretize the range into a LiDAR scene descriptor, $k_l$, and a sonar scene descriptor $k_s$ as appropriate. Each scene descriptor is encoded as 
\vspace{-3mm}
\begin{align}
    k = (f(r_1), ... , f(r_N)), f:r_i \xrightarrow{} \mathbb{R}
\end{align}
\vspace{-1mm}
where the contents of each discrete bin, $f(r_i)$, is the number of occurrences inside that range bin. Lastly, we normalize the counts in each histogram bin by dividing them by the maximum count in each descriptor. A candidate loop closure, consisting of a pair of discrete time steps, $i$ from an AUV and $j$ from the USV, is generated using a KD tree. 

\subsubsection{Point Cloud Registration}
\label{registration}
Given a loop closure candidate, a pair of indexes, the $ith$ AUV step, and the $jth$ USV step, we apply point cloud registration between the two. First, we gather a 2D sonar submap, denoted as $\mathcal{S}_s$, which spans from time step $i$ to $i-w_s$. Where $w_s$ is the user specified sonar window size. The submap is formed as, 
\vspace{-3mm}
\begin{align}
    \mathcal{S}_s = \{\Tbf_{i-0} \mathcal{P}_{s,i-0}, 
                            \Tbf_{i-1} \mathcal{P}_{s,i-1}, 
                            ...,
                            \Tbf_{i-w_s} \mathcal{P}_{s,i-w_s} \},
\end{align}
by applying the rigid body transform, $\Tbf_{i-w_s}$, which denotes the transformation between step $i$, and preceding steps inside the window. We apply the same submap process to the 2D LiDAR point clouds as $\mathcal{S}_L$, denoted as 
\vspace{-3mm}
\begin{align}
    \mathcal{S}_L = \{\Tbf_{j-0} \mathcal{P}_{L ,j-0}, 
                            \Tbf_{j-1} \mathcal{P}_{l,j-1}, 
                            ...,
                            \Tbf_{j-w_s} \mathcal{P}_{l,j-w_s} \},
\end{align}
Each submap is voxel down-sampled to the same resolution cell size, $\mathcal{V}$. Each voxel down-sampled cloud is denoted as $\mathcal{\hat{S}}_s$ and $\mathcal{\hat{S}}_L$. We then use iterative closest point (ICP) \cite{icp} to align the two submaps, specifically Go-ICP \cite{go-icp}. Go-ICP is used to solve for the unknown 3 DoF translation $T$ and rotation, $R$ between the two submaps with \textit{no initial guess.} Standard ICP refines Go-ICP results as a secondary registration step. 
\vspace{-3mm}
\begin{align}
\label{icp_transform}
     T = \begin{pmatrix} \Delta x \ 
    \Delta y\end{pmatrix}^\top, R = \begin{pmatrix} \cos{\Delta \theta}\ -\sin{\Delta \theta} \\  \sin{\Delta \theta} \ \;\;\;\;\cos{\Delta \theta}
    \end{pmatrix}
\end{align}

\textcolor{black}{Go-ICP is used, due to the poor initial overlap between submaps, which is a failure case for standard ICP \cite{icp}.} Once the submaps are aligned, we compute a registration quality metric, overlap, denoted as $O_s$. $O_s$ is the percentage of points in $\mathcal{\hat{S}}_s$ that have a nearest neighbor within 1-meter. This simple metric is used to check for outliers, by only retaining loop closures with an $O_s$ greater than $O_{min}$.

\subsubsection{Robust Outlier Rejection}
\label{pcm-section}
If an inter-robot loop closure has an $O_s >= O_{min}$, then it is further verified by a robust outlier rejection system, Pairwise Consistent Measurement Set Maximization algorithm (PCM) \cite{pcm}. PCM analyzes loop closures as a group, performing geometric verification and retaining only the largest set of consistent inter-robot loop closures. 
\begin{align}
    \label{pcm_con}
    \norm{( \mathbf T_{L_l}^{L_j} \cdot \mathbf T_{L_j}^{s_i} \cdot \mathbf T_{s_i}^{s_k}) \cdot {\mathbf T_{L_l}^{s_k}}^{-1} }^2_2<\epsilon.
\end{align}
A pair of loop closures is consistent if they satisfy the condition in Equation \ref{pcm_con}. Place recognitions between LiDAR and sonar are denoted as $\mathbf T_{L_j}^{s_i}$ and $\mathbf T_{L_l}^{s_k}$. $i, j, k, l$ are the discrete time steps in each data form, with each represented by LiDAR, $L$, and sonar, $s$. The transformation between each data type are denoted by $ \mathbf T_{L_l}^{L_j}$ and $\mathbf T_{s_i}^{s_k}$.

\subsection{Centralized Multi-robot Pose Graph}
The centralized multi-robot factor graph is formulated onboard the USV. The USV's pose chain from Section \ref{lidar-pose-graph} is added as sequential factors between poses, denoted as $\mathbf f^{\text{LO}}$. Next, each AUV's pose chain from Section \ref{auv-pose-graph} is added as sequential factors between poses, denoted as $\mathbf f^{\text{DO}}$. Lastly, each inter-robot loop closure is added as $\mathbf f^{\text{IR}}$, completing the pose graph in Equation \ref{multi-robot-graph}.
\begin{flalign}
\label{multi-robot-graph}
\mathbf f(\boldsymbol \Theta) = \mathbf  f^{\text{0}}(\boldsymbol \Theta_0) & \prod_i \mathbf f^{\text{LO}}_{i}(\boldsymbol \Theta_i) \prod_j \mathbf f^{\text{DO}}_j(\boldsymbol \Theta_j) \prod_q \mathbf f^{\text{IR}}_q(\boldsymbol \Theta_q)
\end{flalign}
This graph solves all robot poses jointly, yielding a centralized understanding of the environment. Pose information provides enhanced AUV state estimates and maps of the environment, both underwater and above-water, in a common reference frame. Lastly, these position estimates serve as an initial guess for all subsequent applications of point cloud registration in Section \ref{registration}.

\section{Experiments}
\subsection{Hardware Overview}
In this work, we use the heavily modified ClearPath Kingfisher shown in Figure \ref{fig:kingfisher} to collect datasets for validation. Kingfisher is equipped with a custom flight control system based on the BlueCube ecosystem. For inertial navigation, the Kingfisher uses a Nortek DVL-1000, Vectornav VN100 IMU, and Bluerobotics Bar02 depth sensor. On the perception side, we utilize a VLP-32C, a 32-beam above-water LiDAR, and, mounted on an underwater pole, a Blueprint Subsea Oculus M750d multi-beam imaging sonar. For dataset ground-truthing, we use the Here3+ RTK-GPS system. 

\subsection{Datasets}
When collecting datasets, we operate our Kingfisher in two different modalities. Option one is to collect USV data: above-water LiDAR and IMU. Option two is to collect AUV data: IMU, DVL, and underwater imaging sonar. This method enables us to collect multiple datasets with RTK-GPS ground truth and replay them concurrently with simulated communications to validate the proposed multi-robot system using real-world perceptual data. 

We consider three real-world environments, each with a multi-robot team of a single USV and two AUVs. Each environment includes robots with different starting locations and trajectories. \textbf{Bridge} is shown in Figure \ref{fig:bridge_sat_image}, consisting of several large seawalls and several sizable bridge pylons. \textbf{Waterfront} is shown in Figure \ref{fig:waterfront_sat_image}, consisting of seawalls, several piers, and a large footbridge supported by small-diameter pier pilings. \textbf{Harbor} is shown in Figure \ref{fig:harbor_sat_image}, consisting of seawalls, piers, floating docks, pier pilings, and three ships. Additionally, all environments feature a small number of moving objects, including pedestrians, cars, and boats, as well as substantial magnetic interference from steel-hulled ships and underwater high-voltage power lines. These three environments represent different, complex littoral settings to validate this work. 

\subsection{Metrics}
We use two primary metrics to evaluate results: error and time. For SLAM error, we compare estimated trajectories to RTK-GPS data to develop mean-absolute error (MAE) and root-mean-squared error (RMSE) to capture both the average and size of error distributions. When considering time, we evaluate module runtime in milliseconds (ms).

\subsection{Multi-robot SLAM Ablation Study}
In this section, we compare varying versions of the proposed system to a single-agent underwater SLAM system for robots with imaging sonar \cite{Wang-2021}. We consider the error of each AUV in the multi-robot system separately and compare it to the same dataset using a single-agent SLAM system.

In our ablation study, we vary the sonar window size and the addition of the PCM robust outlier rejection block. We use a fixed LiDAR window size of three, as the LiDAR sensor range of 200 meters is significantly larger than the sonar range of 30 meters. Due to their comparatively small sensor range, sonar submaps are prone to creating outliers because of their limited context. Therefore, we evaluate the following cases: \textbf {Case 1}: Small sonar window size, without PCM, \textbf{Case 2}: Small sonar window size, with PCM, \textbf{Case 3}: Large sonar window size, without PCM and \textbf{Case 4}: Large sonar window size, with PCM.

\begin{table}[]
\begin{tabular}{lllll}
\midrule
 Case &  AUV ID & MAE (meters) &  RMSE (meters) \\
 \midrule
 \multirow{2}{*}{Bridge Case 1} &  1 & 15.6 & 18.02\\
                           &  2 & 2.12 & 2.33\\
 \hdashline
 \multirow{2}{*}{Bridge Case 2} &  1 & Fail & Fail \\
                           &  2 & 5.25 & 5.97\\
 \hdashline
 \multirow{2}{*}{Bridge Case 3} &  1& 15.16 & 18.02\\
                           &  2&  \textbf{2.12} & \textbf{2.33} \\
 \hdashline
 \multirow{2}{*}{Bridge Case 4} &  1& \textbf{2.80} & \textbf{3.37} \\
                           &  2&  2.13 & 2.39\\
 \hdashline
 Bridge Single             &  1&  3.35 & 3.75 \\
 Robot SLAM                &  2&  2.80 & 3.23 \\
 \midrule
 \multirow{2}{*}{Waterfront Case 1} &  1 & Fail & Fail \\
                           &  2& Fail & Fail\\
 \hdashline
 \multirow{2}{*}{Waterfront Case 2} &  1 & 23.95 & 28.64 \\
                           &  2& 24.1 & 30.58 \\
 \hdashline
 \multirow{2}{*}{Waterfront Case 3} &  1 & 14.08 & 20.14\\
                           &  2& \textbf{4.13} & \textbf{5.57} \\
 \hdashline
 \multirow{2}{*}{Waterfront Case 4} &  1 & \textbf{9.38} & \textbf{12.06} \\
                           &  2 & 4.46 & 4.96 \\
 \hdashline
 Waterfront Single             &  1 & 13.92 & 18.85 \\
 Robot SLAM                &  2&  7.22 & 7.80 \\
 \midrule
 \multirow{2}{*}{Harbor Case 1} &  1& Fail & Fail\\
                           &  2&  Fail & Fail\\
 \hdashline
\multirow{2}{*}{Harbor Case 2} &  1& Fail & Fail\\
                           &  2&  Fail & Fail\\
 \hdashline
 \multirow{2}{*}{Harbor Case 3} &  1& \textbf{4.3} & \textbf{5.31}\\
                           &  2& 3.83 & \textbf{4.43}\\
 \hdashline
 \multirow{2}{*}{Harbor Case 4} &  1& 5.44 & 7.02\\
                           &  2& \textbf{3.7} & 4.47 \\
 \hdashline
 Harbor Single             &  1& 5.92 & 7.44 \\
 Robot SLAM                &  2& 8.92 & 11.44\\
 \midrule

\end{tabular}
\caption{\textbf{Real world multi-robot SLAM results.} MAE/RMSE is reported for each of the two AUVs in our three robot team. Bridge results are shown at the top, waterfront in the middle and harbor at the bottom. We report data for the ablation study, cases 1-4 and a baseline comparison to single robot SLAM with each AUV dataset. }

\label{slam-table}
\vspace{-6mm}
\end{table}

\begin{figure*}[th]
\centering
\subfigure[Bridge AUV 1\label{fig:bridge_auv_1}]{\includegraphics[height=3.2cm]{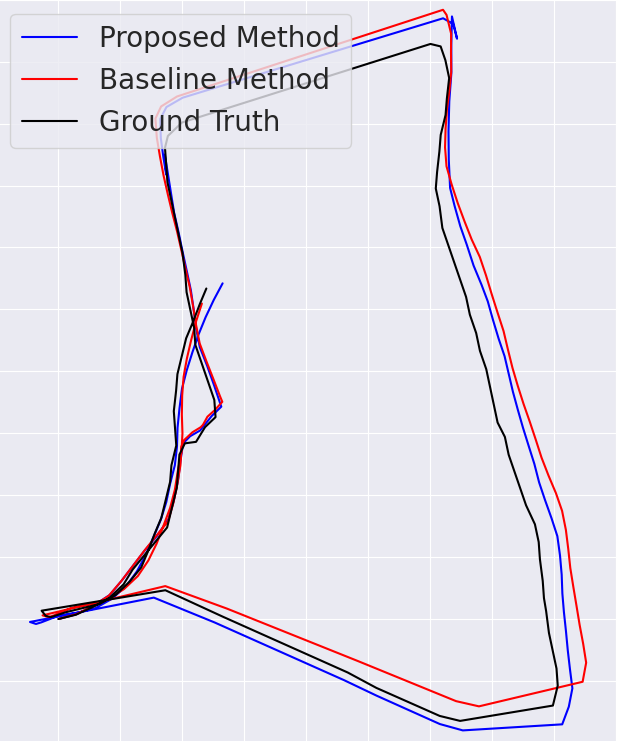}}\ \;
\subfigure[Bridge AUV 2\label{fig:bridge_auv_2}]{\includegraphics[height=3.2cm]{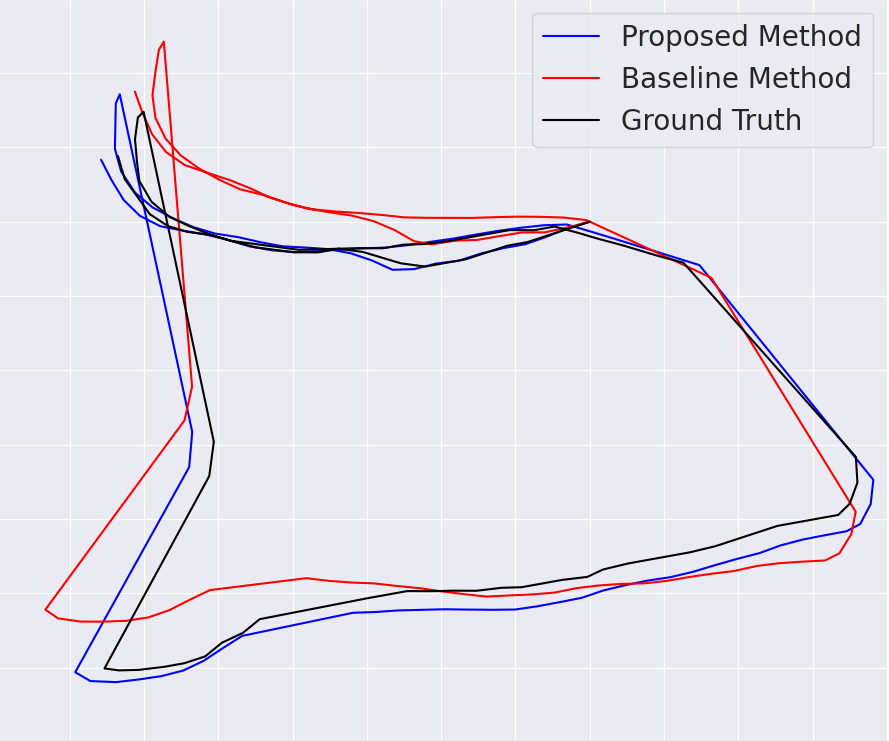}}\ \;
\subfigure[Waterfront AUV 1\label{fig:Waterfront_auv_1}]{\includegraphics[height=3.2cm]{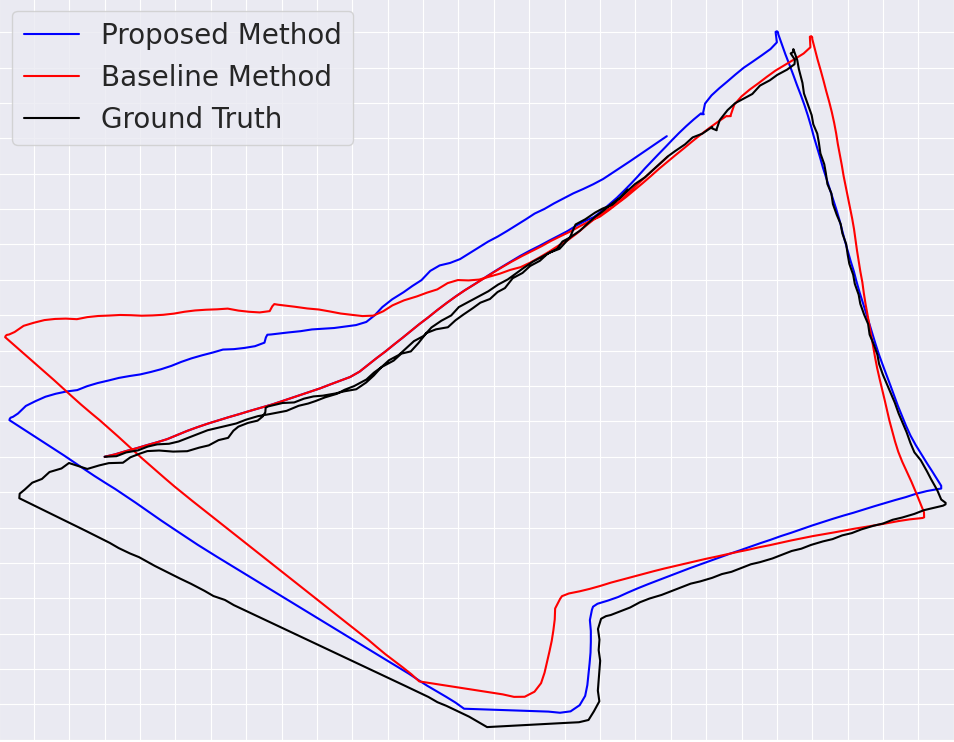}}\ \;
\subfigure[Waterfront AUV 2\label{fig:Waterfront_auv_2}]{\includegraphics[height=3.2cm]{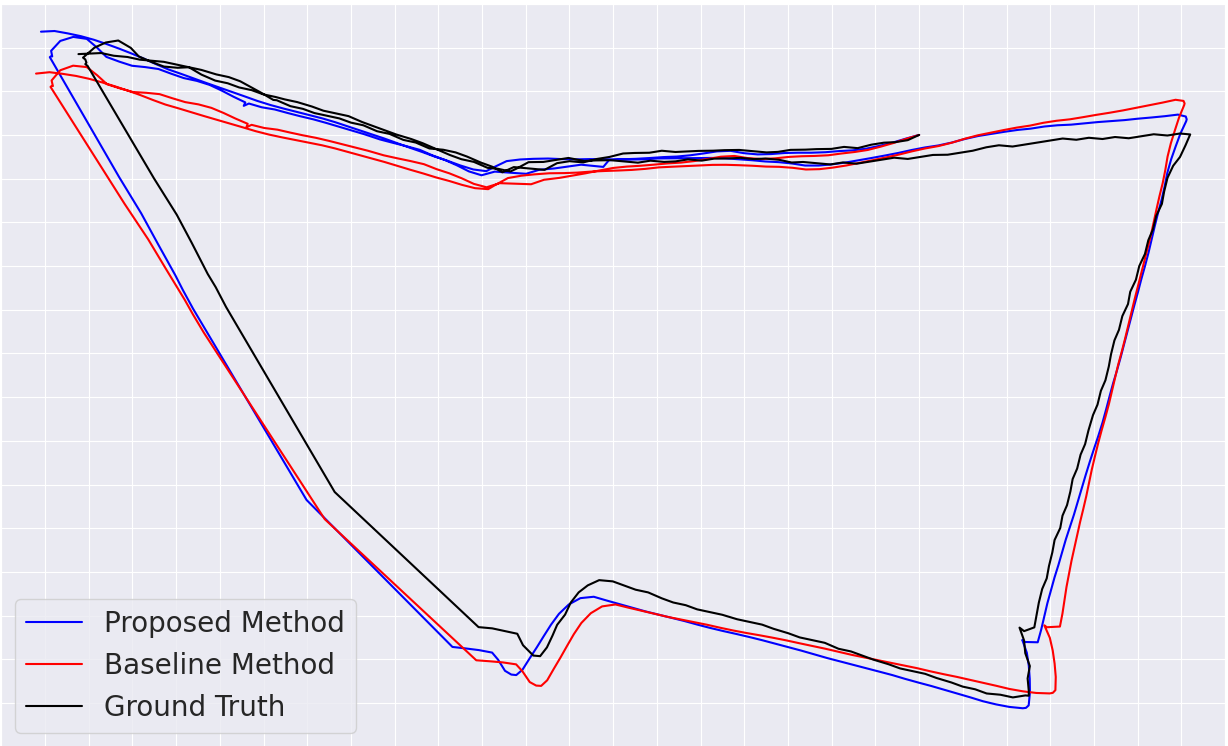}}\\
\subfigure[Harbor AUV 1\label{fig:Harbor_auv_1}]{\includegraphics[height=3.2cm]{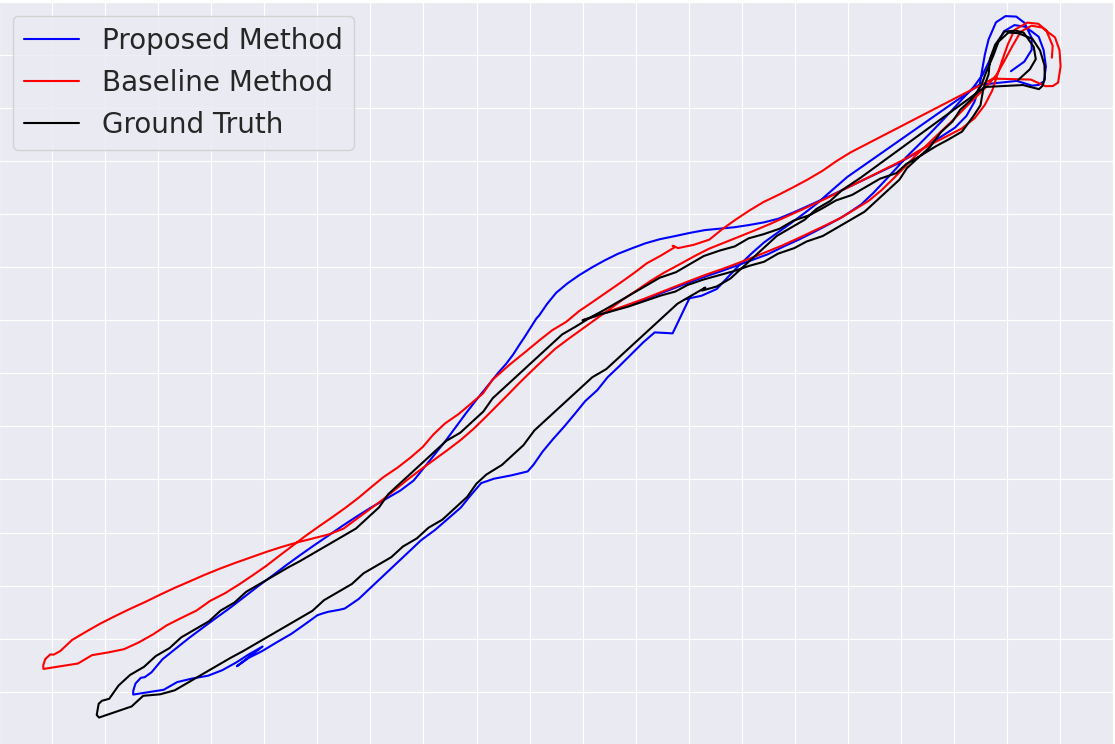}}\ \;
\subfigure[Harbor AUV 2\label{fig:Harbor_auv_2}]{\includegraphics[height=3.2cm]{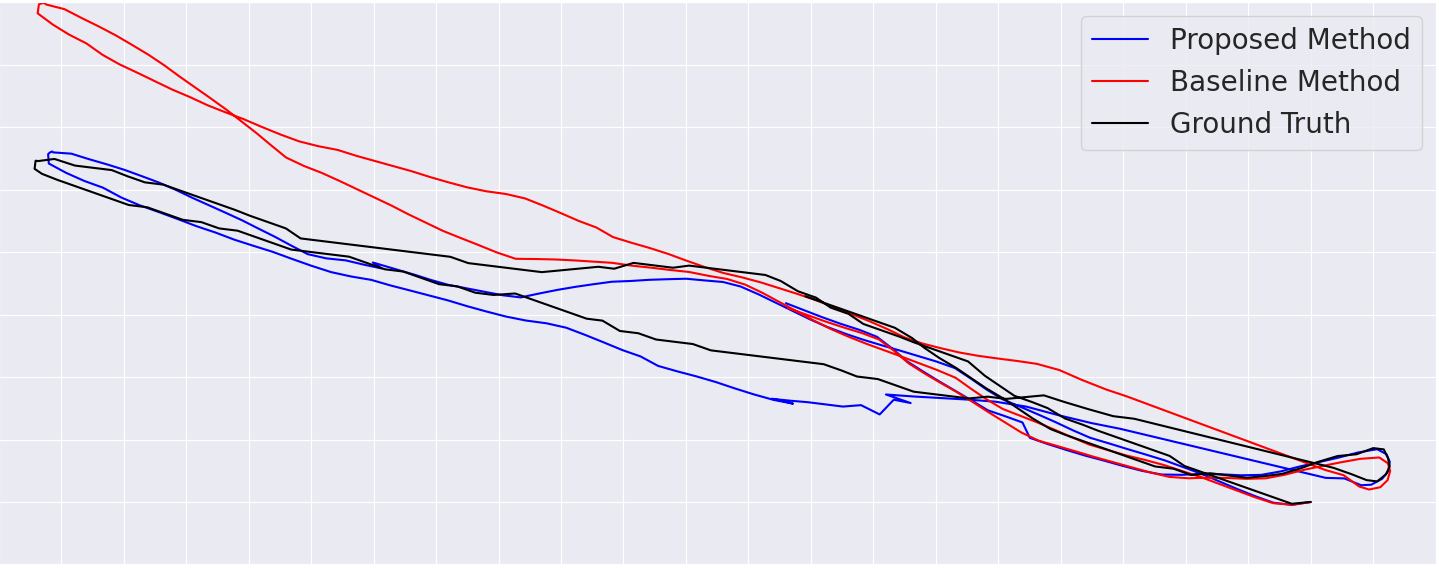}}\ \;

\caption{\textbf{Qualitative Results.} Each plot shows a single AUV in the system. Black lines show the ground truth trajectory from RTK-GPS. The red lines indicate the baseline trajectory from single-robot underwater SLAM. Blue lines show the trajectory from the proposed method, centralized multi-robot SLAM. Results from bridge are shown in (a) and (b). Results from waterfront are shown in (c) and (d). Results from harbor are shown in (e) and (f). Grid lines are shown at a 10-meter cell size for all plots. }
\label{fig:envs}
\vspace{-5mm}
\end{figure*}

\subsubsection{Bridge Results}
In this section, we consider the bridge dataset with a small window size of 3 and a large window size of 32. Quantitative results are shown at the top of Table \ref{slam-table}, and qualitative results are shown in Figures \ref{fig:bridge_auv_1} and \ref{fig:bridge_auv_2}. For case 1 of the ablation study (window size 3, no PCM), there is a high error for AUV 1. Case 2 (window size 3, with PCM) shows a failed SLAM result for AUV-2 due to outliers. Case 3 (window size 32, no PCM) yields results similar to those of Case 1. Lastly, in case 4 (window size 32 with PCM), there is a low error for both AUVs in the system. The ablation study shows that using a large window and PCM for the bridge environment is well-founded, attenuating any outliers. Moreover, when comparing case 4 to the baseline, single-robot SLAM, case 4 shows improvements in both MAE and RMSE. 

\subsubsection{Waterfront Results}
In this section, we consider the waterfront dataset with a small window size of 3 and a large window size of 50. Quantitative results are shown in the middle of Table \ref{slam-table}, and qualitative results are shown in Figure \ref{fig:Waterfront_auv_1} and \ref{fig:Waterfront_auv_2}. Case 1 (window size 3, no PCM) exhibits failure for both AUVs, while Case 2 (window size 3, with PCM) displays significant errors. Cases 3 (window size 50, no PCM) and 4 (window size 50, with PCM) improve these errors. Again, using a large window, both with and without PCM, shows the best MAE/RMSE in the ablation study compared to single-robot SLAM. Any improvement without PCM is most likely because of PCM's extremely judicious nature. PCM can, on occasion, be overly strict when accepting loop closures, resulting in the false rejection of helpful information.  

\subsubsection{Harbor Results}
In this section, we consider the harbor dataset with a small window size of 3 and a large window size of 30. Quantitative results are shown at the bottom of Table \ref{slam-table}, and qualitative results are shown in Figure \ref{fig:Harbor_auv_1} and \ref{fig:Harbor_auv_2}. Case 1 (window size 3, no PCM) exhibits failure for both AUVs, while Case 2 (window size 3, with PCM) also reports failure. In case 3 (window size 35, no PCM), this error is attenuated. In case 4 (window size 35, with PCM), again, PCM is a stringent outlier rejection system, which can create higher MAE/RMSE. Similar to the waterfront, there are multiple meter improvements in MAE and RMSE compared to single robot SLAM.

\subsubsection{SLAM Results Summary}
For all three environments, there is an improvement when using a larger window size, with mixed results when adding PCM. However, PCM is an important addition due to its robustness against outliers, which is crucial for a SLAM mission to succeed. When comparing the proposed system to the single-robot SLAM baseline, significant improvements are observed in all three environments. Lastly, an example composite map, including sonar and LiDAR from different trajectories, is shown in Figure \ref{fig:lead_off}.

\begin{table}[b]
\begin{tabular}{lllll}
\midrule
 Data Type &  Mean Size & STD Size & Min Size & Max Size \\
 \midrule
 Raw Sonar Image & 1978.37 & 1978.37 & 1978.37 & 1978.37\\
 Jpeg Sonar Image & 75.14 & 15.73 & .07 & 124.0 \\
 CFAR Coordinates & 14.28 & 15.5 & .10 & 79.14\\
 CFAR Compressed & 5.63 & 4.76 & .06 & 23.55\\
 \textcolor{black}{DRACo-SLAM \cite{draco}} & \textcolor{black}{1.12} & \textcolor{black}{1.11} & \textcolor{black}{.02} & \textcolor{black}{6.336} \\

\end{tabular}
\caption{\textbf{Perceptual Message Sizes (k-bits).} The first row reports the cost to transmit a sonar image, the second row is the same, with jpeg compression applied. The third row is only transmitting CFAR detection coordinates, and the fourth row is with rectangle compression applied. The last row shows DRACo-SLAM compression \cite{draco}. } 
\label{data-size-table}
\vspace{-3mm}
\end{table}

\subsection{Perceptual Data Analysis}

While many complexities are associated with using a system like this, we focus on data rate, noting COTS acoustic modems have rates as high as 62.5 k-bits/second \cite{evo}. We evaluate the size of perceptual messages in k-bits in Table \ref{data-size-table} across all three environments. We compare raw sonar images, JPEG, pixels selected by CFAR in Section \ref{sonar-cfar}, and the rectangle compression in Section \ref{sonar-compression}. First, we observe the costly nature of exchanging raw and JPEG images. \textcolor{black}{Next, we compare to DRACo-SLAM \cite{draco} compression, which reduces data resolution from 0.05 to 0.25 meters, and as shown in \cite{draco}, has a negative impact on SLAM error. } Lastly, we observe the savings associated with CFAR down-sampling and the further savings with rectangle compression. \textcolor{black}{Network statistics for bandwidth usage are provided as average kbits/second: Bridge 0.95 kbits/second, Waterfront 1.86 kbits/second, and Harbor 3.38 kbits/second.} These results are encouraging. From a data rate perspective, perceptual data may be exchanged over acoustic communications. 

\begin{table}[b]
\begin{tabular}{lll}
\midrule
 Module &  Mean Runtime (ms) & STD Runtime (ms)  \\
 \midrule
PCM & 6.05 & 5.23  \\
Point Cloud Registration & 766.11 & 254.60 \\
Rectangle Compression & 115.64 & 40.0 \\

\end{tabular}
\caption{\textbf{Runtime statistics in ms.} Point cloud registration is from Section \ref{registration}, PCM is the outlier rejection system in Section \ref{pcm-section}, and rectangle compression is the sonar data compression system in Section \ref{sonar-compression}. } 
\label{run-time-table}
\vspace{-5mm}
\end{table}

\subsection{Runtime Analysis}
In this section, we consider the runtime of modules in the proposed system. Runtime results are summarized in Table \ref{run-time-table} using an AMD Ryzen 9 9950X 16-Core Processor.

We report the mean and standard deviation of runtime for the point cloud registration system in Section \ref{registration}, PCM in Section \ref{pcm-section}, and rectangle compression in Section \ref{sonar-compression}. Firstly, rectangle compression and PCM report low runtimes, with point cloud registration taking more time. It is essential to evaluate these runtime statistics in the context of how often each module is queried. Recall that the proposed method operates over a discrete set of keyframes. Given the vehicle speed for our datasets, these are added at a mean rate of 5469.53 ms with a standard deviation of 1920.61 ms, indicating the total average runtime of the modules in Table \ref{run-time-table} is lower than the rate at which they are required.

\section{Conclusions}
In this work, we propose a multi-robot SLAM system for USVs equipped with 3D LiDAR and AUVs with imaging sonar. This centralized multi-robot system demonstrates improved AUV state estimation compared to single AUV SLAM. Moreover, this system enables multi-robot mapping of complex littoral environments, with structures above and below the water surface. We performed validation in three real-world environments, using a team of three robots consisting of a single mother-ship USV and two AUVs. \textcolor{black}{When considering failure modes, robots must actually observe the same structures for loop closure to be possible. Additionally, perceptual aliasing is persistent, with many structures appearing identical in sonar imagery, necessitating a strict outlier rejection system. However, this can lead to numerous missed opportunities, resulting in higher underwater vehicle error. }

This system represents an exciting development, marking the first multi-robot SLAM system to incorporate USVs and AUVs through indirect encounters. Future work will leverage real acoustic communication systems, both to validate the system in its current form in the field and to add acoustic range measurements between all robots in the system. 


{}

\end{document}